\documentclass{article}

\usepackage{arxiv}
\usepackage[utf8]{inputenc} 
\usepackage[T1]{fontenc}    
\usepackage{hyperref}       
\usepackage{url}            
\usepackage{booktabs}       
\usepackage{amsfonts}       
\usepackage{nicefrac}       
\usepackage{microtype}      
\usepackage{lipsum}		
\usepackage{graphicx}
\usepackage{natbib}
\usepackage{doi}

\usepackage{algorithm}
\usepackage{algpseudocode}
\usepackage{amsbsy}
\usepackage{amsmath}
\usepackage{amssymb}
\usepackage{balance}
\usepackage{graphicx}
\usepackage{multirow}
\usepackage{subcaption}

\title{Learning Pruned Structure and Weights Simultaneously from Scratch: an Attention based Approach}
\author{
    Qisheng He,
    Weisong Shi,
    Ming Dong\\
    Computer Science\\
    Wayne State University \\
}

\usepackage{bibentry}

\begin{document}

\maketitle

\begin{abstract}
As a deep learning model typically contains millions of trainable weights, there has been a growing demand for a more efficient network structure with reduced storage space and improved run-time efficiency. Pruning is one of the most popular network compression techniques. In this paper, we propose a novel unstructured pruning pipeline, Attention-based Simultaneous sparse structure and Weight Learning (ASWL). Unlike traditional channel-wise or weight-wise attention mechanism, ASWL proposed an efficient algorithm to calculate the pruning ratio through layer-wise attention for each layer, and both weights for the dense network and the sparse network are tracked so that the pruned structure is simultaneously learned from randomly initialized weights. Our experiments on MNIST, Cifar10, and ImageNet show that ASWL achieves superior pruning results in terms of accuracy, pruning ratio and operating efficiency when compared with state-of-the-art network pruning methods.
\end{abstract}

\section{Introduction}
\label{sec.Introduction}
Deep learning models such as Convolutional Neural Network (CNN) have achieved great success in various computer vision tasks such as image recognition \cite{ImageNet, ResNet}, object detection \cite{FasterRCNN, MaskRCNN}, and semantic segmentation \cite{SemanticSegmentation}. A key avenue for deploying deep learning models is a mobile device or an edge server in order to reduce latency and ensure data privacy for users. As a deep learning model typically contains millions of trainable weights, this practice has been accompanied by a growing demand for a more efficient network structure with reduced storage space and improved run-time operating efficiency.

Recently, there has been a resurgence in neural network compression techniques (e.g., pruning and quantization) \cite{EDropout}. One can compress a given neural network architecture into an extremely small size without compromising on the model performance. Even efficient networks \cite{MobileNets} can be further compressed even though they already have a small footprint. It was shown, for instance, that SqueezeNet \cite{SqueezeNet} can be compressed to 0.5MB, $510\times$ smaller than AlexNet \cite{AlexNet}. 

\begin{figure}[ht!]
	\centering
	\includegraphics[width=\columnwidth]{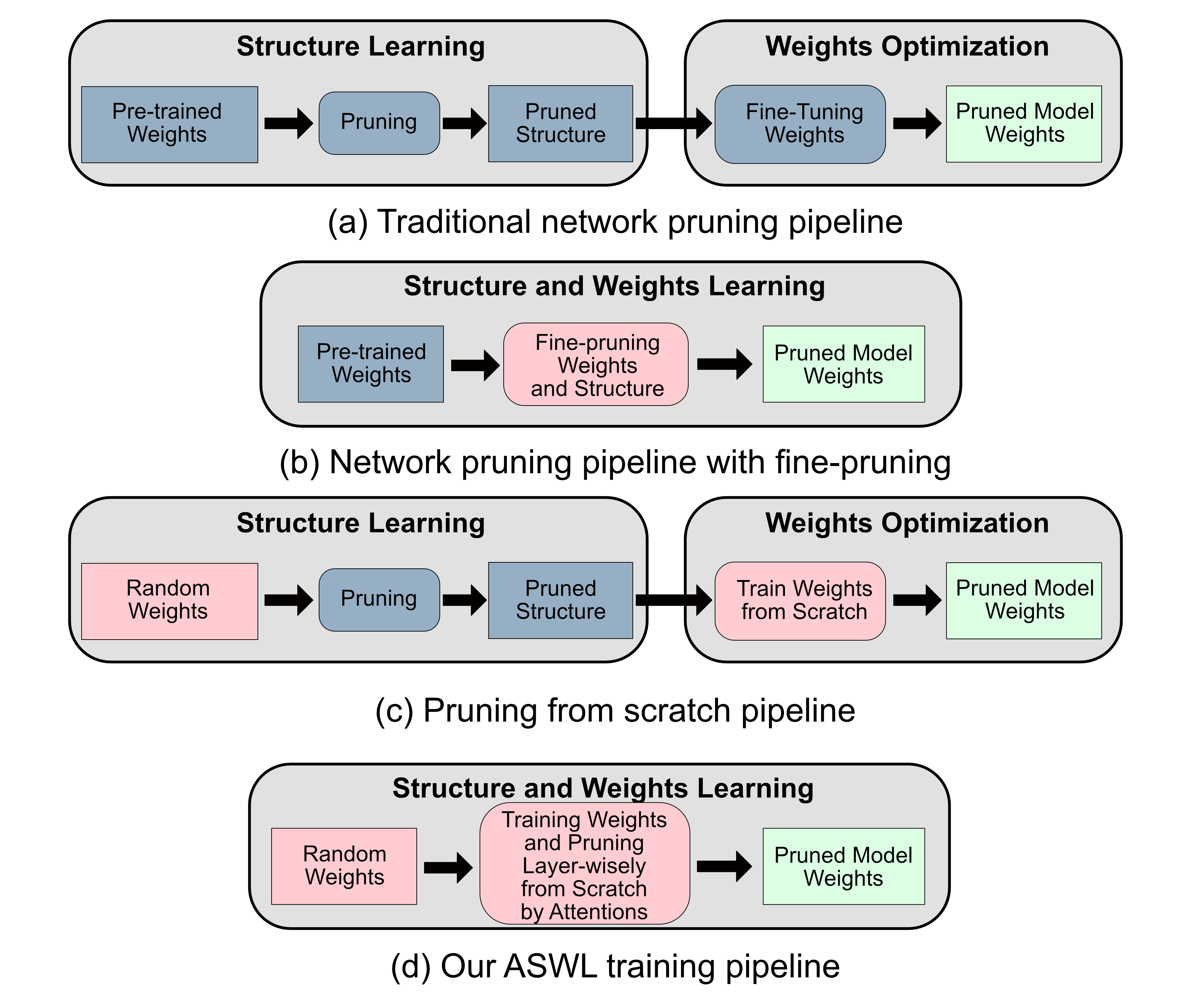}
	\caption{Comparison of neural network pruning pipelines. (a) Traditional network pruning learns pruned structure and fine-tunes from a pre-trained model. (b) Network fine-pruning learns both structure and weights jointly from a pre-trained model. (c) Pruning from scratch first generates a pruned structure and then trains the pruned model from randomly initialized weights. (d) ASWL learns both sparse structure and weights simultaneously through layer-wise attention from scratch in a unified procedure.}
	\label{fig.PipelineComparison}
\end{figure}

Pruning, which removes connections or neurons in a network, is one of the most popular network compression techniques. It reduces both the size and the computational complexity of a model since pruned weights do not need to be calculated or stored. Moreover, pruning is size-efficient as deep sparse models consistently outperform shallow dense models with almost no loss of or even a greater accuracy \cite{ToPruneOrNotToPrune}. When network pruning was first introduced for compression, redundant connections were pruned using a three-step pipeline illustrated in Fig. \ref{fig.PipelineComparison}(a): first, the network is trained to learn which connections are important; next, the unimportant connections are pruned; finally, the network is retrained to fine-tune the weights of the remaining connections \cite{LearningBothWeightsAndConnections}. The same three-step pipeline was later employed to perform layer-wise pruning in a layer-wise optimal brain surgeon \cite{LayerWiseOptimalBrainSurgeon}.

It was shown that structure learning and weights optimization could be combined to simplify the pruning process as Fig \ref{fig.PipelineComparison}(b) shows. Fine-pruning \cite{FinePruning} proposed a principled method in which the pruning ratio of each layer is first predicted by the Bayesian optimization with a pre-trained network, and then the network is jointly fine-tuned and compressed with the predicted pruning ratios. The joint training was frequently used in quantization methods known as the Straight-Through Estimator (STE) \cite{STEForQuantization} to avoid zero gradients during back-propagation. The advantage of such a pipeline is that it allows the pruning status to adapt during the change of weights \cite{ClipQ}. However, pruning with pre-training is a time-consuming procedure. One needs to repeat pruning and fine-tuning several times to find a good sparse configuration \cite{SparseNetworksFromScratch}. In the meantime, using a reward function to find the pruning ratio for each layer is also an expensive procedure \cite{NeuralEpitomeSearch}.

Later work, therefore, shows that the weights of a model and its pruned structure can be directly learned from randomly initialized weights (Fig. \ref{fig.PipelineComparison}(c)). Pruning from Scratch \cite{PruningFromScratch} proposed a novel network pruning pipeline that first learns the pruned structure directly from randomly initialized weights and then optimizes the weights of the pruned network. This kind of pipeline bypasses the time-consuming pre-training procedure. Nevertheless, the structure learning and weights optimization proceed separately here so that it only prunes weights based on fixed regularization before training the weights \cite{AutoCompress}.

Recently, the attention mechanism was brought into network pruning and achieved great success \cite{PCAS, GlobalRanking}. In this paper, we propose a new pruning pipeline, Attention-based Simultaneous sparse structure and weight Learning (ASWL). As shown in Fig. \ref{fig.PipelineComparison}(d), in ASWL, the layer-wise sparsity and weights are jointly learned from scratch in a unified training pipeline. Specifically, we first use the attention mechanism to learn the importance of each layer in a network and determine the corresponding pruning ratio. Then, we jointly prune the layer under the guidance of the pruning ratio and update the unpruned weights. The major contributions of our work are summarized as follows:
\begin{itemize}
\item ASWL provides a unified framework that combines both \emph{layer-wise pruning} and weights optimization to learn a pruned network from randomly initialized weights.
During ASWL training, both weights of the dense network and the sparse network are tracked so that the pruning ratio for each layer is simultaneously learned as the weights change.
\item In ASWL, layer-wise pruning decisions are made through a novel attention-based approach. Attention scales in the attention mechanism are set across each layer instead of previous channel-wise or weight-wise attention mechanism. A pruning ratio is then directly computed based on the learned attention value for each layer.
\item ASWL no longer needs the time-consuming pre-training procedure in network pruning but provides equivalent pruning results. Through extensive experiments on benchmark datasets, we demonstrate that ASWL leads to superior pruning results in terms of accuracy, network size, and operating efficiency when compared with state-of-the-art pruning methods.
\end{itemize}

\begin{figure*}
	\centering
	\includegraphics[width=4in]{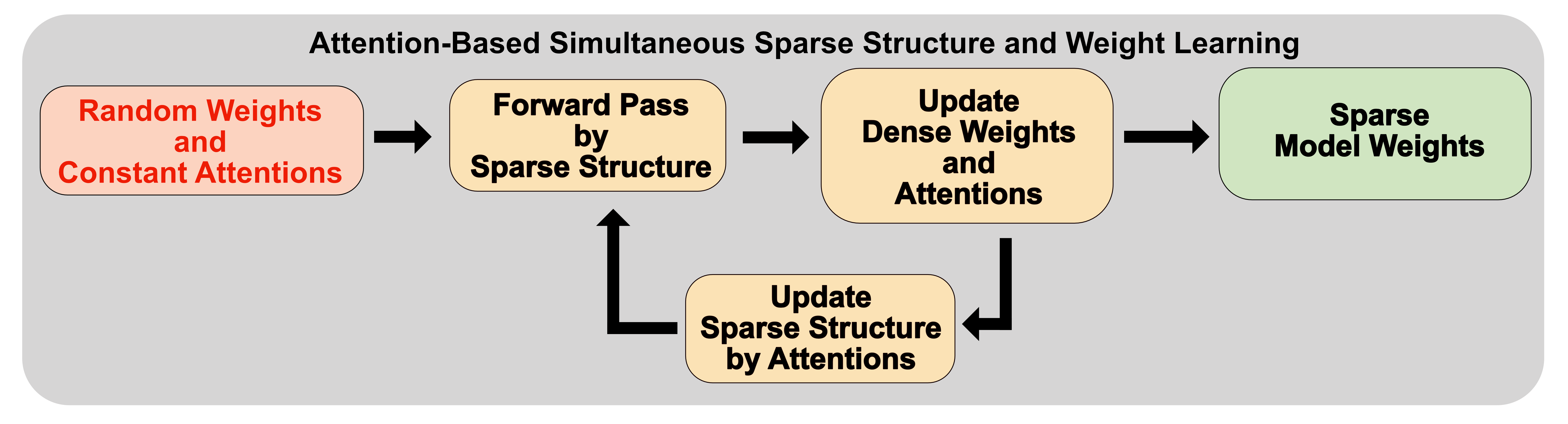}
	\caption{ASWL simultaneous training procedure with the attention mechanism.}
	\label{fig.DetailedPipeline}
\end{figure*}

\section{Related Work}
\label{sec.RelatedWorks}
Neural network pruning techniques were used to avoid the overfitting problem in early work. For example, optimal brain surgeon \cite{OptimalBrainSurgeon} and dynamic sparsely connected artificial neural networks \cite{PhonemeProbabilityEstimation} pruned those connections that lead to a limited change of validation errors. The Dropout was introduced to prune or remove neurons randomly with a given ratio so that the complexity of the model is reduced. Later, pruning methods were employed to identify and remove unnecessary weights to reduce the size as well as the run-time latency of a targeted model. While some of the pruning methods like ThiNet \cite{ThiNet} and Stripe-wise Pruning \cite{PruningFilterInFilter} prune feature maps or filter stripes, most others remove connections/neurons because sparse structures provided by unstructured pruning typically lead to a higher accuracy with a lower redundancy of weights \cite{AutoCompress}.

When pruning methods were first introduced to deep compression, they removed all connections whose weights were lower than a threshold \cite{LearningBothWeightsAndConnections}. It was shown that pruning unimportant weights results in almost no loss of accuracy as the weights in deep neural networks are highly redundant. Then, methods were proposed to identify unnecessary weights layer-wisely \cite{EnergyAwearePruning, LayerWiseOptimalBrainSurgeon, NISP}, which offers flexible pruning ratios on different layers and thus achieves better results. Automated Machine Learning (AutoML) was also introduced into pruning \cite{AMC}, where the design space was efficiently sampled using reinforcement learning to find the best-pruned structure of a pre-trained model. Additionally, the attention mechanism was recently employed to find unimportant weights \cite{AttentionBasedPruning, Shift, GAL}. Most of them introduced attention scales across each channel or each weight of the targeted layer. For example, LeGR-based pruning \cite{GlobalRanking} trained pairs of parameters (scale and shift) across each channel and then pruned output features accordingly.

More recently, the common three-step pruning pipeline, i.e., pre-training, pruning and fine-tuning, has been simplified. Fine-pruning \cite{FinePruning} jointly fine-tunes and compresses the pre-trained network with the pruning ratio of each layer predicted by the Bayesian optimization. Later, the lottery ticket hypothesis stated that dense randomly-initialized neural networks contain subnetworks, known as winning tickets, that reach accuracy equivalent to its original network with a similar number of training iterations \cite{LotteryTicketHypothesis}. The hypothesis was mathematically proved in \cite{ProvingLotteryTicket}. Based on such findings, Pruning from Scratch \cite{PruningFromScratch} proposed to prune networks from randomly initialized weights. Specifically, it first learns the channel importance for each layer with a sparsity regularizer, then searches for a pruned structure with a method adopted from Network Slimming \cite{NetworkSlimming} but used floating-point operations per second (FLOPS) as constraints. The pruned structure is then optimized from random weights. However, sanity-checking \cite{SanityChecking} showed that initial tickets are hard to be learned from the training data. Thus, it is necessary to keep the pruned structure flexible during training. To this end, more recent work trained sparse networks from scratch by regrowing some connections after a fixed number of iterations (e.g., per epoch) based on the gradient momentum, aka, sparse momentum, of the weights in these iterations \cite{SparseNetworksFromScratch}. Nonetheless, when one of the connections converges during training, its sparse momentum will also trend to zero, often leading to a sub-optimal structure and weights during the training. 

\section{Attention-based Simultaneous Learning}
\label{sec.Pipeline}
The objective of our new training pipeline, Attention-based Simultaneous sparse structure and weight Learning (ASWL), is to train both pruned structure and model weights from randomly initialized weights. Different from existing pruning methods, ASWL obtains layer-wise pruned deep learning models without pre-training. As shown in Fig. \ref{fig.DetailedPipeline}, we first convert a model into an attention-based model by replacing its traditional weighted layers with attention-based weighted layers. Instead of learning the pruned structure by regrowing connections after a fixed number of iterations, we pruned the weights layer-wise based on the pruning ratio calculated from the learned attention of the target layer in every iteration. In the mean time, the unpruned parameters are tracked and learned directly from the forward pass of the pruned structure. Such simultaneous learning tracks both pruned and unpruned weights during training and thus gives an adaptive pruning decision on the sparse structure. 

\subsection{Attention-based Neural Networks}
The sparsity of a pruned network is weakly related to its output and hence is hardly to be learned directly \cite{DynamicModelPruning}. Recent research shows that the attention mechanism provides a promising solution \cite{PCAS, AttentionBasedPruning}. Unlike the traditional methods which set attention scalars channel-wisely or weight-wisely, we introduced a much simpler layer-wise attention mechanism, where an attention scalar value was defined across all $L$ layers in our attention-based neural network. Such model is denoted as $f(x; \mathbf{w}, \mathbf{A})$ where $x$ is the input of the model, $\mathbf{w}$ is the traditional trainable weights, and $\mathbf{A} = \{a_1, a_2, ..., a_L\}$ is the layer-wise attention values. Given a scalar gate value $a_l \in (0,1]$ for the $l^{th}$ layer as the attention, it is multiplied with the output of the layer. That is, assuming the output of the $l^{th}$ original layer is $f_l(x_{l-1}; \mathbf{w}_l)$, the attended output is modified as $\hat{f}_l(x_{l-1}; \mathbf{w}_l, a_l) = a_l \cdot f_l(x_{l-1}; \mathbf{w}_l)$. Each attention value is initialized to 0.5 at the beginning. During training, both weights and attentions are updated. With pruning applied, the optimization objective of an attention-based neural network with compressed weights $\mathbf{\hat{w}}$ is:
\begin{equation}
	\label{eq.optimization}
	\underset{\mathbf{\hat{w}}, \mathbf{A}}{min} \sum_i^N {\mathfrak{L}(f(x_i; \mathbf{\hat{w}}, \mathbf{A}))} + \gamma \Psi(\mathbf{A}) + \lambda \sum_j \mathbf{\hat{w}}_j^2
\end{equation}
where $\mathfrak{L}(\cdot)$ denotes the cross-entropy loss, $\Psi(\cdot)$ is the sparsity regularizer for structure learning (discussed in the next section), L2 regularizer encourages all weights to be small \cite{L2Regularizer}, and $\gamma$ and $\lambda$ are the coefficients for the sparsity regularizer and L2 regularizer, respectively.

\subsection{Pruning Ratios and Sparsity Regularizer}
\label{sec.Pipeline.Regularizer}
Layer-wise pruning methods typically require a pre-trained model as the starting point to search for a pruned structure. Recent work in \cite{FinePruning} and \cite{AMC} use naive Bayesian to optimize the pruning ratio layer-wisely in a given network. However, employing this kind of optimization in every training iteration is prohibitively time-consuming. In order to facilitate simultaneous optimization of structure and weights in each iteration, we propose a more efficient algorithm to calculate the pruning ratios directly from the attentions in our attention-based model.

\begin{algorithm}[htbp]
	\caption{Training Algorithm for Attention-based Simultaneous sparse structure and Weight Learning}	
	\label{alg.Training}
	\begin{algorithmic}[1]
		\Function{Train}{$iterations$, $\alpha$}
			\State $\mathbf{w}, \mathbf{A} \leftarrow$ initialize() 
			\\
			\For {$i$ from 1 to $iterations$} 
				\State $\mathbf{p} \leftarrow (1 - \mathbf{A})^{\alpha}$ 
				\State $\mathbf{\hat{w}} \leftarrow prune(\mathbf{w}, \mathbf{p})$ 
				\\
				\State $\mathbf{x} \leftarrow getNextBatch()$
				\State $\mathfrak{l} \leftarrow \sum_i^N {\mathfrak{L}(f(x_i; \mathbf{\hat{w}}, \mathbf{A}))} + \gamma \Psi(\mathbf{A}) + \lambda \sum_i \mathbf{\hat{w}}_i^2$ 
				\\
				\State $\mathbf{G} \leftarrow calculateGradients(\mathfrak{l}; \mathbf{\hat{w}}, \mathbf{A})$ 
				\State $\mathbf{w}, \mathbf{A} \leftarrow updateWeights(\mathbf{G}; \mathbf{w}, \mathbf{A})$ 
			\EndFor
		\EndFunction
	\end{algorithmic}
\end{algorithm}

In \cite{PruningFromScratch, AttentionBasedPruning}, it was shown that scaling  the network weights will suppress the unimportant ones, resulting in a pruning effect. As a result, the learned attentions can be used to represent the importance of each layer. That is, if a layer has a larger attention, it is considered to be more important so that it needs to be pruned less, and vice versa. Here, we introduce a positive hyper-parameter, the pruning factor $\alpha$, to gain more control when computing the pruning ratio $p_l$ based on the given attention $a_l$ of the $l^{th}$ layer:
\begin{equation}
	\label{eq.pruningRatio}
	p_l(a_l) = (1 - a_l)^\alpha,
\end{equation}
where $a_l$ is the attention for the $l^{th}$ layer and $\alpha$ is the pruning factor. To ensure not all weights are pruned, we limited the maximum pruning ratio to be $99\%$. Since our attentions are applied along the layer dimension when weights are pruned, the overall sparsity $S$ of our attention-based model is computed as:
\begin{equation}
	\label{eq.sparsity}
	S(\mathbf{A}) = \frac{\sum_l (1 - p_l(a_l)) n_{\mathbf{w}_l}}{\sum_l n_{\mathbf{w}_l}},
\end{equation}
where $p_l$ is the pruning ratio and $n_{\mathbf{w}_l}$ is the total number of unpruned weights in the $l^{th}$ layer. Note that $n_{\mathbf{w}_l}$ is a constant as it is pre-determined for each layer.

Regularizers like L1 or L2 encourage the network weights $\mathbf{w}$ to be small, but not necessarily zeros. Thus, an additional regularizer is required in ASWL to encourage the pruning procedure to remove unimportant weights in each layer. Moreover, the L1/L2 regularizer encourages attentions to be zeroes without giving any consideration on the sparsity of the entire model, especially when the pruning factor is not set to 1. As a result, we adopted the sparsity regularizer proposed in \cite{PruningFromScratch} as follows and combined it with L2 in ASWL:
\begin{equation}
	\label{eq.attentionRegularizer}
	\Psi(\mathbf{A}) = S(\mathbf{A})^2
\end{equation}
The square sparsity regularizer is differentiable \cite{AutoPruner} and will help minimize the layer-wise sparsity during optimization.

\subsection{Simultaneous Sparse Structure Learning and Weight Optimization}
Traditional pruning methods learn the pruned structure first, and then optimize the weights based on the pruned structure. The pruned structure can be easily found if a pre-trained model is available. However, with the weights training from random initialized values, efficient weights might change gradually from shallow to deep layers \cite{SparseNetworksFromScratch}. Meanwhile, a dense neural network contains a subnetwork that has the same accuracy even without fine-tuning \cite{ProvingLotteryTicket}. These results motivated us to perform simultaneous learning on both the sparse structure and weights in ASWL by tracking both the dense network (through the backward propagation) and the sparse subnetwork (through the forward pass). 

In ASWL training, both weights for the dense network and the sparse network are tracked so that the structure is simultaneously learned as the weights change. The parameters of the sparse network are pruned by removing at least the bottom $p_i$ percent dense weights in the $i_{th}$ layer based on the absolute weight values $|w_i|$. After training, the attention values are applied to the weights and discarded together with the weights in the dense network. Only the weights in the sparse network are stored. 

Each training iteration in ASWL contains four steps: (1) the network classification loss is calculated through a forward pass based on the compressed weights $\mathbf{\hat{w}}$ and attentions $\mathbf{A}$, (2) both uncompressed weights $\mathbf{w}$ and attentions $\mathbf{A}$ are updated through back-propagation, (3) the pruning ratio $\mathbf{p}$ is computed by the new attentions $\mathbf{A}$ and the pruning factor $\alpha$, and (4) the model is compressed layer-wise using the pruning ratio $\mathbf{p}$, and the compressed weights $\mathbf{\hat{w}}$ are updated. The detailed training procedure is summarized in Algorithm \ref{alg.Training}. In ASWL, through simultaneous optimization, weights that have been pruned at first may be recovered later, and the weights defined as important at first can be pruned, all depending on the evolution of the network structure.

\section{Experimental Results}
\label{sec.Results}
In this section, we perform extensive experiments with ASWL on VGG-16 \cite{VGG}, ResNet50 \cite{ResNet}, and MobileNetV2 \cite{MobileNetV2}. For VGG-16 and ResNet50, we simply replace the traditional convolutional layer and dense layer with our attention-based convolutional layer and dense layer. For MobileNetV2, we replace the $1 \times 1$ point-wise convolutional layer with an attention-based convolutional layer but left the depth-wise layer uncompressed since 99\% of the parameters and calculations are contained in point-wise convolutional layers. There are totally three hyperparameters in ASWL: the coefficients of sparsity and L2 regularizer ($\gamma$ and $\lambda$), and the pruning factor $\alpha$. They are specified later in different experiments.

Our ASWL training pipeline was implemented in TensorFlow \cite{TensorFlow}. All models are trained on a computer with Intel i7 8700K CPU, 16GB RAM, and two NVIDIA RTX 2080 Ti graphic cards, each of which has 11GB of GDDR SDRAM. \textbf{The source code of this work is available in the supplemental materials} with detailed comments and will be made publicly available after the review period of AAAI 2022.

\begin{figure}[ht!]
	\centering
	\begin{subfigure}{0.9\columnwidth}
		\includegraphics[width=0.9\columnwidth]{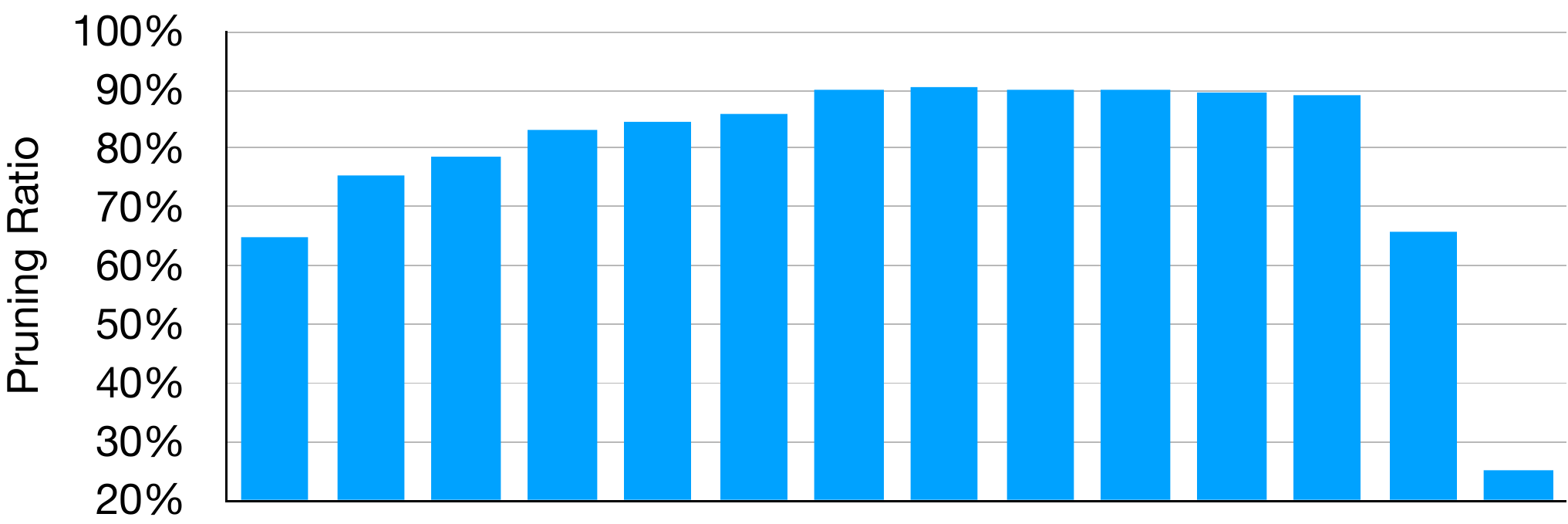}
		\caption{Pruning details for VGG-16 with $\alpha=1.5$}
	\end{subfigure}
	\begin{subfigure}{0.9\columnwidth}
		\includegraphics[width=0.9\columnwidth]{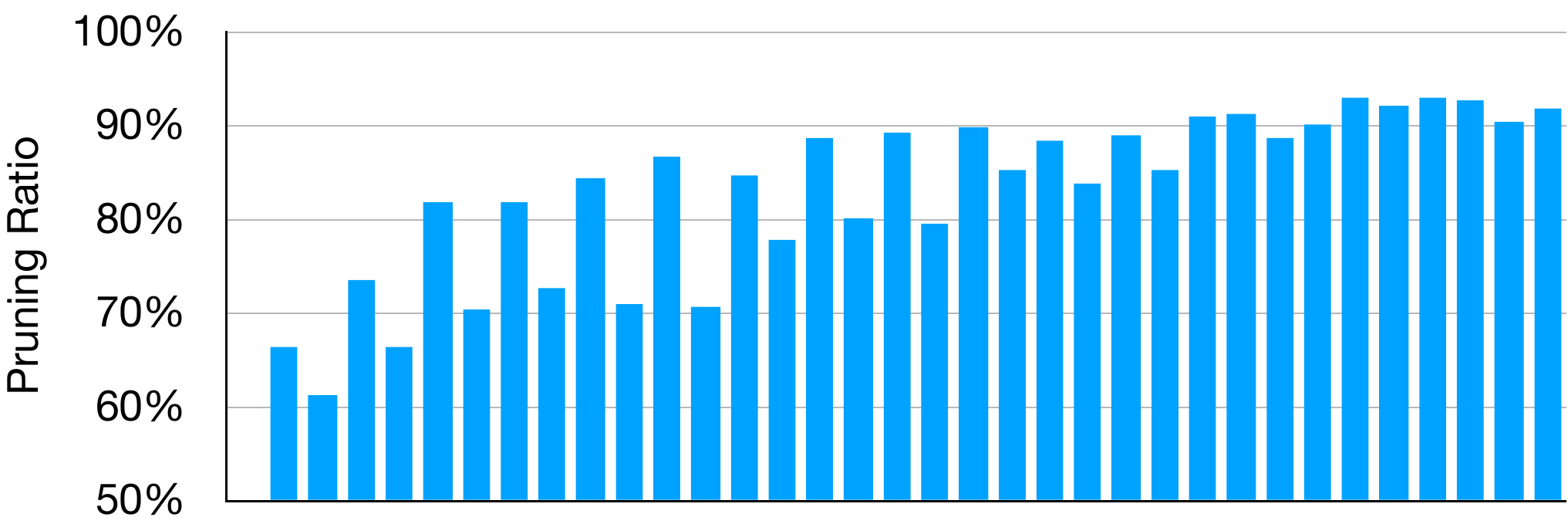}
		\caption{Pruning details for MobileNetV2 with $\alpha=1$}
	\end{subfigure}
	\begin{subfigure}{0.9\columnwidth}
		\includegraphics[width=0.9\columnwidth]{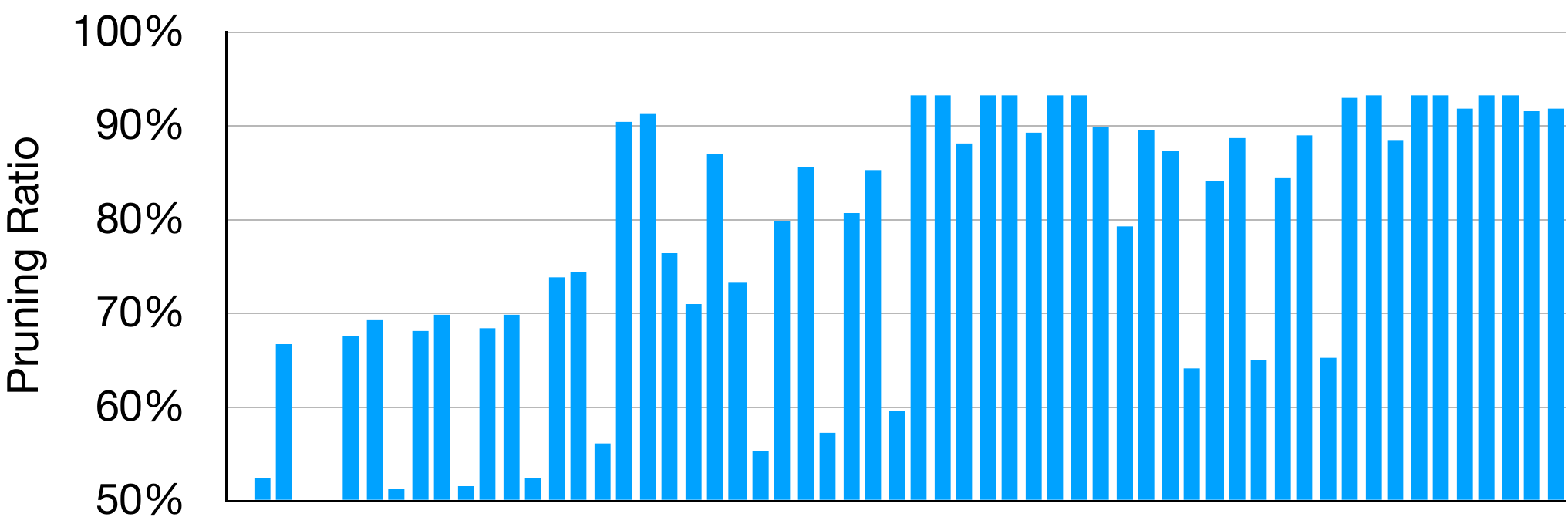}
		\caption{Pruning details for ResNet56 with $\alpha=1$}
	\end{subfigure}
	\caption{Layer-wise pruning details on MNIST for (a) VGG-16 with $\alpha=1.5$, where the last two layers are attention-based dense layers, and all others are attention-based convolutional layers, (b) MobileNetV2 with $\alpha=1$, where the first layer is an attention-based convolutional layer with a $3 \times 3$ filter, and others are point-wise ($1 \times 1$ filter) attention-based convolutional layers, (c) ResNet56 with $\alpha=1$, where the last layer is an attention-based dense layer, and others are attention-based convolutional layer.}
	\label{fig.MNISTPruningDetails} 
\end{figure}

\begin{figure}[ht!]
	\centering
	\begin{subfigure}{0.9\columnwidth}
		\includegraphics[width=0.9\columnwidth]{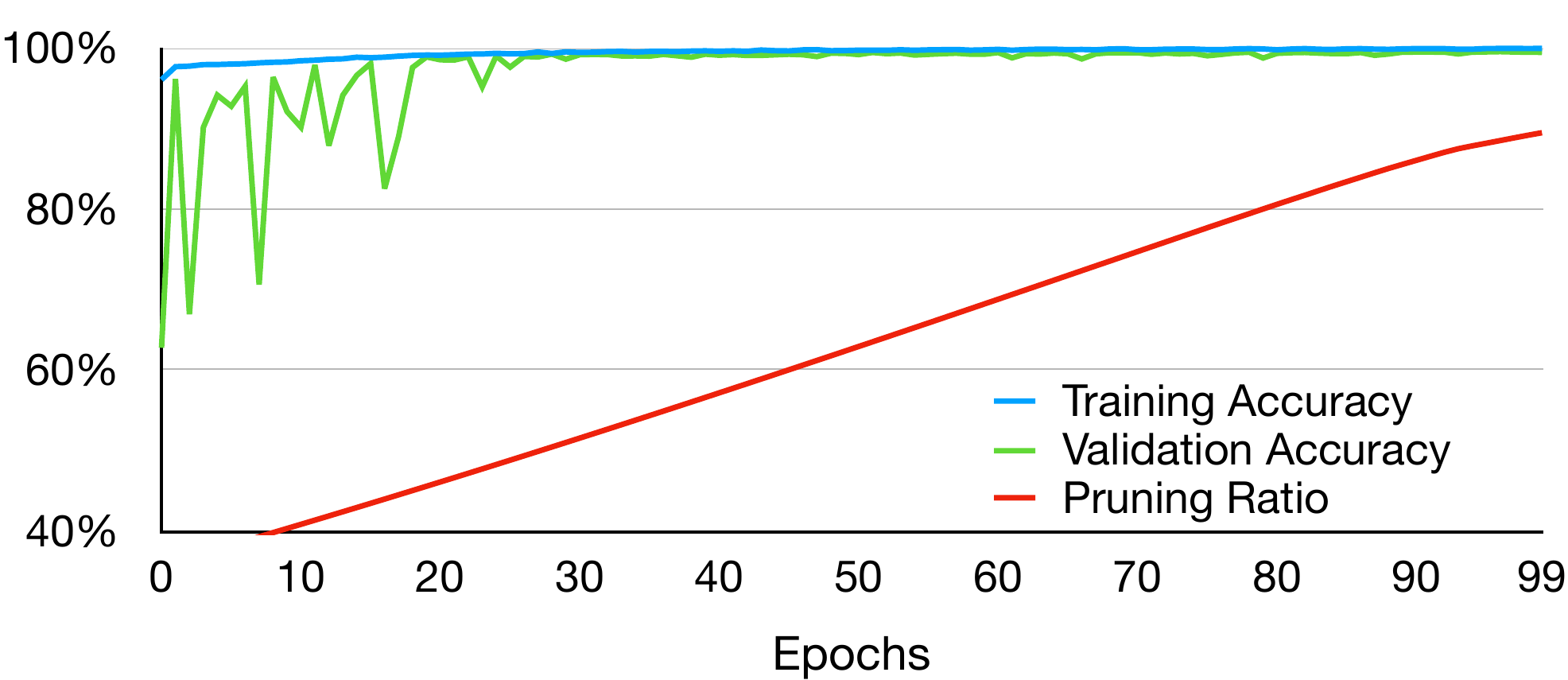}
		\caption{Training graph for VGG16 with $\alpha=1.5$}
	\end{subfigure}
	\begin{subfigure}{0.9\columnwidth}
		\includegraphics[width=0.9\columnwidth]{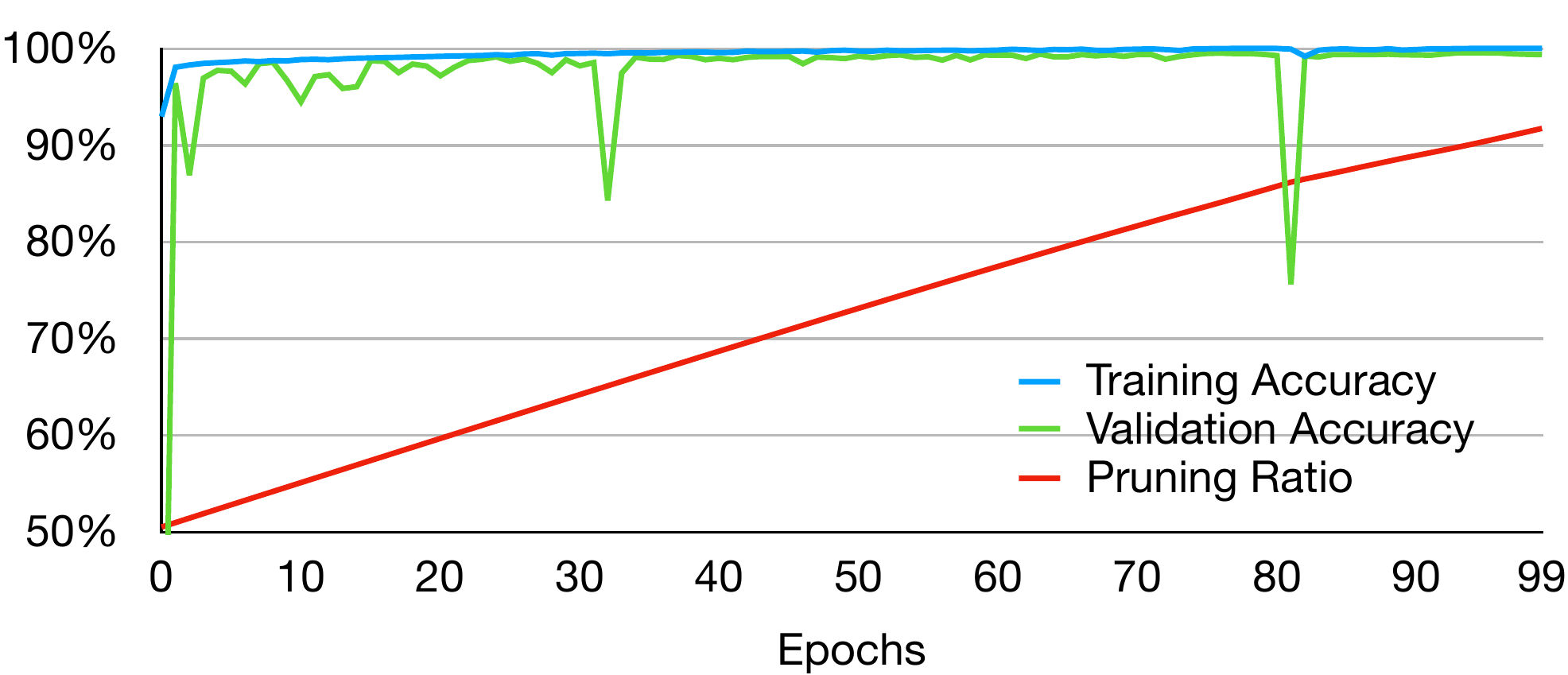}
		\caption{Training graph for MobileNetV2 with $\alpha=1$}
	\end{subfigure}
	\begin{subfigure}{0.9\columnwidth}
		\includegraphics[width=0.9\columnwidth]{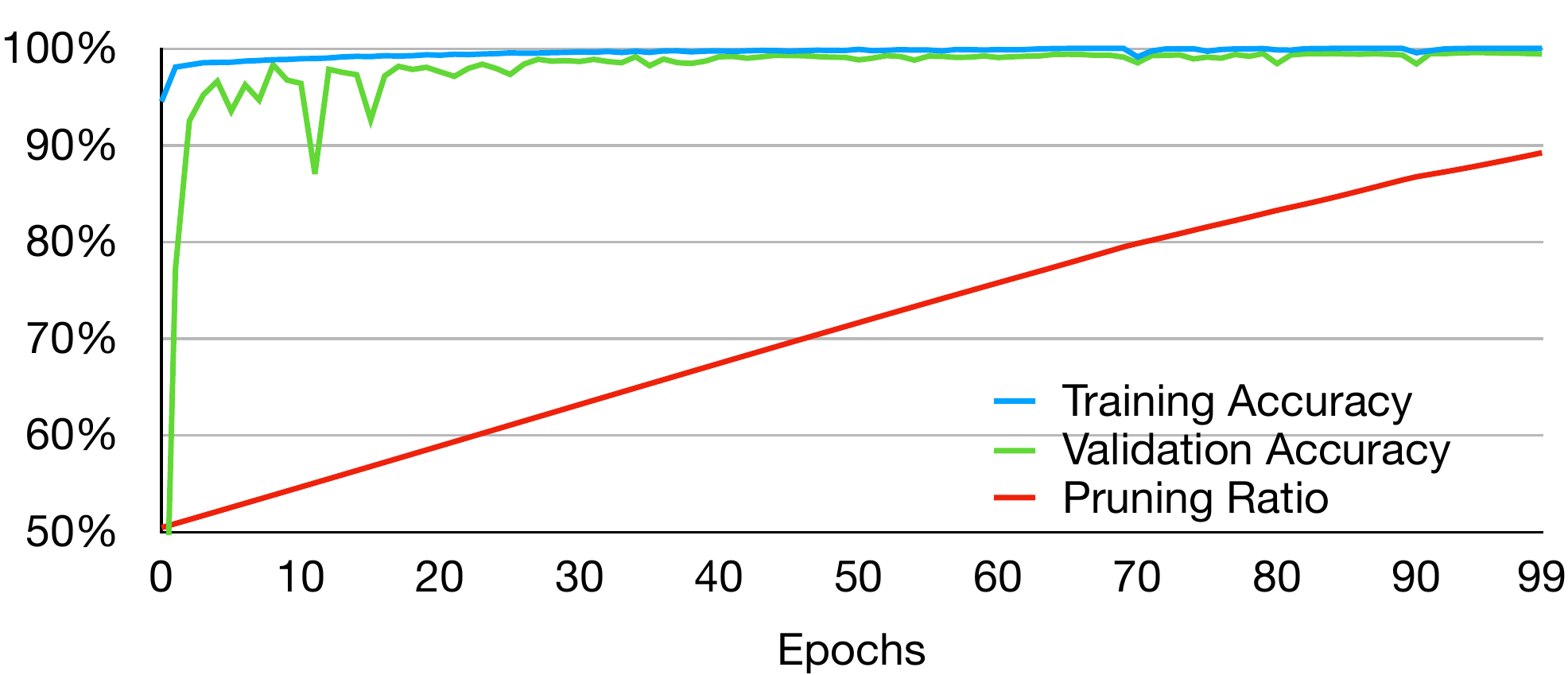}
		\caption{Training graph for ResNet56 with $\alpha=1$}
	\end{subfigure}
	\caption{MNIST Training graph: training accuracy (blue line), validation accuracy (green line), and pruning ratio (red line) for (a) VGG-16 with $\alpha=1.5$, (b) MobileNetV2 with $\alpha=1$, and (c) ResNet56 with $\alpha=1$.}
	\label{fig.MNISTTrainingGraph}
\end{figure}

Specifically, we conducted our experiments on the standard MNIST \cite{MNIST}, Cifar-10 \cite{Cifar10} and ImageNet (ILSVRC-2012 in version 2.0.1) \cite{ImageNet} datasets and compared with the following pruning methods: Learning both Weights and Connections for Efficient Neural Networks (ENN) \cite{LearningBothWeightsAndConnections}, Rethingking Network Pruning (RNP) \cite{RethinkingNetworkPruning}, Discrimination-aware Channel Pruning (DCP) \cite{DisAwareChannelPruning}, Convolutional neural network pruning with structural redundancy reduction (SRR) \cite{SRR}, Network Pruning via Performance Maximization (NPPM) \cite{NPPM}, Learned Threshold Pruning (LTP) \cite{LearnedThresholdPruning}, Pruning from Scratch (PFS) \cite{PruningFromScratch}, Dynamic Sparse Reparameterization (DS) \cite{DynamicSparse}, Pruning Filter in Filter (PFF) \cite{PruningFilterInFilter}, WoodFisher (WF) \cite{WoodFisher}, Sparse Networks from Scratch (SNS) \cite{SparseNetworksFromScratch}, and Dynamic Pruning with Feedback (DPF) \cite{DynamicModelPruning}.

\begin{table*}
	\begin{center}
		\begin{tabular}{c| c | c | c c }
			\hline
			\multirow{2}{*}{\textbf{Model}} & \textbf{Pruning Factor} & \textbf{Baseline} & \textbf{$\Delta$ Accuracy} & \textbf{Pruning Ratio} \\
			& \textbf{$\alpha$} & \textbf{$\%$} & \textbf{$\%$} & \textbf{$\%$} \\
			\hline
			VGG16-1 & 1 & \multirow{3}{*}{99.51\%} & +0.04\% & 90.69\% \\
			VGG16-1.5 & 1.5 & & +0.09\% & 88.56\% \\
			VGG16-2 & 2 & & +0.03\% & 87.19\% \\
			\hline
			MobileNetV2-1 & 1 & \multirow{3}{*}{99.56\%} & +0.00\% & 90.10\% \\
			MobileNetV2-1.5 & 1.5 & & -0.07\% & 86.32\% \\
			MobileNetV2-2 & 2 & & -0.09\% & 82.03\% \\			
			\hline
			ResNet56-1 & 1 & \multirow{3}{*}{99.50\%} & +0.07\% & 87.74\% \\
			ResNet56-1.5 & 1.5 & & +0.00\% & 83.70\% \\
			ResNet56-2 & 2 & & +0.05\% & 75.61\% \\	
			\hline
		\end{tabular}
		\caption{ASWL training results with different pruning factors on VGG-16 (top row), MobileNetV2 (middle row), and ResNet50 (bottom row) for MNIST.}
		\label{table.MNIST}
	\end{center}
\end{table*}

\begin{table*}
	\begin{center}
		\begin{tabular}{c| c | c | c c }
			\hline
			\multirow{2}{*}{\textbf{Model}} & \textbf{Pruning Factor} & \textbf{Baseline} & \textbf{$\Delta$ Accuracy} & \textbf{Pruning Ratio} \\
			& \textbf{$\alpha$} & \textbf{$\%$} & \textbf{$\%$} & \textbf{$\%$} \\
			\hline
			VGG16-1 & 1 & \multirow{3}{*}{92.64} & +0.50\% & 93.80\% \\
			VGG16-1.5 & 1.5 & & +0.76\% & 91.96\% \\
			VGG16-2 & 2 & & +0.18\% & 78.11\% \\
			\hline
			MobileNetV2-1 & 1 & \multirow{3}{*}{93.14} & +0.27\% & 95.86\% \\
			MobileNetV2-1.5 & 1.5 & & +0.41\% & 94.61\% \\
			MobileNetV2-2 & 2 & & +0.11\% & 78.89\% \\			
			\hline
			ResNet56-1 & 1 & \multirow{3}{*}{93.44} & +0.28\% & 95.97\% \\
			ResNet56-1.5 & 1.5 & & +0.10\% & 94.86\% \\
			ResNet56-2 & 2 & & +0.26\% & 96.19\% \\	
			\hline
		\end{tabular}
		\caption{ASWL training results with different pruning factors on VGG-16 (top row), MobileNetV2 (middle row), and ResNet56 (bottom row) for Cifar10.}
		\label{table.Cifar10}
	\end{center}
\end{table*}

\begin{table*}[ht!]
	\begin{center}
		\begin{tabular}{c | c | c c c | c | c c}
			\hline
			\multirow{2}{*}{\textbf{Model}} & \multirow{2}{*}{\textbf{Method}} & \textbf{Simu.} & \textbf{From} & \textbf{Unstructured} & \textbf{Baseline} & \textbf{$\Delta$ Acc.} & \textbf{Pr. Ratio} \\
			& & \textbf{Training} & \textbf{Scratch} & \textbf{Pruning} & \% & \% & \% \\
			\hline
			\multirow{6}{*}{VGG16}
			& DCP & $\times$ & $\times$ & $\times$ & 93.80\% & +0.17\% & 52.1\% \\
			& RNP & $\times$ & $\times$ & $\checkmark$ & 93.76\% & -0.04\% & 80.0\% \\
			& PFS & $\times$ & $\checkmark$ & $\times$ & 93.44\% & +0.19\% & 93.6\% \\
			& PFF & $\times$ & $\checkmark$ & $\times$ & 93.25 & -0.40\% & 92.7\% \\
			& SNS & $\circ$ & $\checkmark$ & $\checkmark$ & 93.51\% & -0.50\% & 95.0\% \\
			& \textbf{ASWL (Ours)} & $\pmb{\checkmark}$ & $\pmb{\checkmark}$ & $\pmb{\checkmark}$ & 92.81\% & \textbf{+0.76\%} & \textbf{92.0\%}  \\
			\hline
			\multirow{7}{*}{ResNet56} 
			& DCP  & $\times$ & $\times$ & $\times$ & 93.80\% & -0.31\% & 92.8\% \\
			& RNP & $\times$ & $\times$ & $\checkmark$ & 93.80\% & -0.31\% & 93.1\% \\
			& PFS & $\times$ & $\checkmark$ & $\times$ & 93.23\% & -0.18\% & 93.1\% \\
			& PFF & $\times$ & $\checkmark$ & $\times$ & 93.10\% & +0.12\% & 77.7\% \\
			& DPF-90 & $\checkmark$ & $\checkmark$ & $\checkmark$ & 94.51\% & -0.56\% & 90.0\% \\
			& DPF-95 & $\checkmark$ & $\checkmark$ & $\checkmark$ & 94.51\% & -1.26\% & 95.0\% \\
			& \textbf{ASWL (Ours)} & $\pmb{\checkmark}$ & $\pmb{\checkmark}$ & $\pmb{\checkmark}$ & 93.44\% & \textbf{+0.28\%} & \textbf{96.0\%} \\
			\hline
		\end{tabular}
		\caption{ASWL training results on Cifar10, and the comparison with state-of-the-art network pruning methods. In the ``simultaneous training'' column, $\times$ indicates static pruned structure during training, $\circ$ indicates that the network connections are iteratively regrown after a fixed number of iterations (e.g., every epoch), and $\checkmark$ indicates fully simultaneous structure and weight training. In the ``from scratch'' column, $\times$ indicates pruning from pre-trained, and $\checkmark$ indicates pruning from scratch. In the ``unstructured pruning'' column, $\times$ indicates structured pruning, and $\checkmark$ indicates unstructured. For each method, accuracy, and pruning ratio are reported, and the best results are highlighted in \textbf{bold}. The same applies to Table \ref{table.ImageNet.Compare}}.
		\label{table.Cifar10.Compare}
	\end{center}	
\end{table*} 

\subsection{Results on MNIST}
We first trained selected models with ASWL on MNIST, which contains 10 different handwriting digits with 60,000 training images and 10,000 testing images. Each of the models was trained with various pruning factors of 1, 1.5, and 2 for 100 epochs by the Adam optimizer at a learning rate of 0.001 and 0.98 decay for each epoch. The attentions of each layer are initialized at 0.5. The hyper-parameter $\gamma$ (the sparsity regularizer coefficient) is used mainly to achieve a desired pruning ratio and set at 0.5 for all models. Following \cite{VGG}, \cite{ResNet}, and \cite{MobileNetV2}, the other hyper-parameter $\lambda$ (the L2 regularizer coefficient) is set at $5 \cdot 10^{-4}$, $0.0001$, and $0.00004$ for VGG16, ResNet 56, MobileNetV2, respectively. The models with the best results are selected.

Table \ref{table.MNIST} shows the ASWL training results on MNIST with VGG16 (Top), MobileNetV2 (Middle), and ResNet56 (Bottom). In most situations, the pruning ratio progressively increases when we reduce the pruning factor. There is no obvious relation found between model accuracy and pruning factors, while a smaller pruning factor offers a greater pruning ratio. With about $10\%$ to $25\%$ of weights in the original dense models, our ASWL provides similar or many times higher accuracy.

Fig. \ref{fig.MNISTPruningDetails} (a) shows the pruning details of each layer in VGG-16 with a pruning factor of $\alpha=1.5$. Deeper attention-based convolutional layers are pruned more than the shallower ones, while the last two attention-based dense layers have pruning ratios much less than the convolutional layers. Fig. \ref{fig.MNISTPruningDetails} (b) demonstrates the pruning ratio of each regular (non depth-wise) attention-based convolutional layer in MobileNetV2 with a pruning factor of $\alpha=1$. Its first convolutional layer with a filter size of $3 \times 3$ has the least pruning ratio, just above $50\%$. Meanwhile, pruning ratios appear to be greater in deeper convolutional layers, similar to the VGG-16 case. The layer-wise pruning details of ResNet56 are shown in Fig. \ref{fig.MNISTPruningDetails} (c), where the last layer is an attention-based dense layer. ResNet56 has much more layers compared with VGG-16 and MobileNetV2 but still has an obvious trend of a higher pruning ratio for a deeper layer. Fig. \ref{fig.MNISTTrainingGraph} (a) shows the training graph of VGG-16 with a pruning factor of $\alpha=1.5$. Overall, the pruning ratio continues increasing as the model is being trained, and the validation accuracy stabilizes after 30 epochs. We observe the same in the training graphs of MobileNetV2 with $\alpha=1$ and ResNetV2 with $\alpha=1$ (Fig. \ref{fig.MNISTTrainingGraph} (b) and (c)).
\subsection{Results on Cifar10}
Cifar10 is a dataset that contains 10 different classes with 50,000 training images and 10,000 testing images. Similar to our experiment on MNIST, we trained the selected models with various pruning ratios of 1, 1.5, and 2. For VGG16, we used an initial learning rate of 0.1 and multiplied 0.5 for every 20 epochs with the SGD optimizer (momentum 0.9), and trained for 250 epochs at a batch size of 128. For ResNet 56, we followed the same settings in \cite{ResNet}. For MobileNetV2, we trained for 350 epochs on SGD optimizer with an initial learning rate of 0.1, which was divided by 10 after 150 and 250 epochs. Again, the hyper-parameter $\gamma$ (sparsity regularizer coefficient) is used to help us achieve a desired pruning ratio and was set at 2.5, 5 and 0.5 for VGG16, ResNet 56, MobileNetV2, respectively. Following \cite{VGG}, \cite{ResNet}, \cite{MobileNetV2}, the other hyper-parameter $\lambda$ is set at $5 \cdot 10^{-4}$, $0.0001$, and $0.00004$ for VGG16, ResNet 56, and MobileNetV2, respectively. The models with the best results are selected.

Table \ref{table.Cifar10} shows the ASWL training results on Cifar10 with VGG16 (Top), MobileNetV2 (Middle), and ResNet56 (Bottom). Similar to MNIST, the results show that small pruning factors typically lead to higher pruning ratios. Additionally, we observe that ASWL offers an improvement on model accuracy with reduced network size and higher run-time efficiency over uncompressed baselines in all cases. The pruning details of the three models the three models are shown in Fig. 1 in supplemental materials. Similar to the case of MNIST, deeper layers are pruned more than the shallower ones except for the last dense layer for classification. Similar training graphs (Fig. 2 in supplemental materials) of the three models were observed as in MNIST that the model converges after about 30 epochs and reaches the highest validation accuracy at around 130 epochs, at which time the overall pruning ratio stops increasing as well. 

Table \ref{table.Cifar10.Compare} compares ASWL with state-of-the-art pruning methods on Cifar10 with ResNet56 (top) and VGG16 (bottom). Clearly, ASWL achieved a higher accuracy than the baseline on both models. Overall, our ASWL model achieved an outstanding pruning ratio with the highest increase of accuracy when compared to the baseline. These results clearly demonstrate the advantages of simultaneous training and layer-wise attention-based pruning in ASWL.

\begin{table*}
	\begin{center}
		\begin{tabular}{c | c c c | c | c c}
			\hline
			\multirow{2}{*}{\textbf{Method}} & \textbf{Simu.} & \textbf{From} & \textbf{Unstructured} & \textbf{Baseline} & \textbf{T1 Acc.} & \textbf{Pr. Ratio} \\
			& \textbf{Training} & \textbf{Scratch} & \textbf{Pruning} & \% & \% & \% \\
			\hline
			ENN & $\times$ & $\times$ & $\checkmark$ & \multirow{11}{*}{76.1\%} & 76.1\% & 60.0\% \\
			CP 0.5$\times$ & $\times$ & $\times$ & $\times$ & & 75.4\% & 48.5\% \\
			SRR & $\times$ & $\times$ & $\times$ & & 75.1\% & 55.1\% (FLOPS) \\
			PFS 0.75$\times$ & $\times$ & $\checkmark$ & $\times$ & & 75.6\% & 63.9\% \\
			DS-20 & $\circ$ & $\checkmark$ & $\checkmark$ & & 73.3\% & 80.0\% \\
			SNS-20 & $\circ$ & $\checkmark$ & $\checkmark$ & & 73.8\% & 80.0\% \\
			NPPM & $\checkmark$ & $\checkmark$ & $\times$ & & 76.0\% & 56.0\% (FLOPS) \\
			LTP & $\checkmark$ & $\times$ & $\checkmark$ & & 73.3\% & 85.6\% \\
			WF-90 & $\checkmark$ & $\times$ & $\checkmark$ & & 75.2\% & 90.0\% \\
			DPF-80 & $\checkmark$ & $\checkmark$ & $\checkmark$ & & 75.5\% & 73.5\% \\
			DPF-90 & $\checkmark$ & $\checkmark$ & $\checkmark$ & & 74.6\% & 82.6\% \\
			\textbf{ASWL (Ours)} & $\pmb{\checkmark}$ & $\pmb{\checkmark}$ & $\pmb{\checkmark}$ & & \textbf{76.5\%} & \textbf{86.1\%} \\
			\hline
		\end{tabular}
		\caption{ASWL training result for ResNet-50 on ImageNet, and the comparison with state-of-the-art network pruning methods. Pruning ratio with FLOPS indicates the pruning ratio of the target method is calculated by FLOPS instead of number of weights.}
		\label{table.ImageNet.Compare}
	\end{center}
\end{table*}

\subsection{Results on ImageNet}
Compared with MNIST and Cifar10, ImageNet is a much larger dataset that contains 1000 classes with 1.2M training images and 50K testing images. We trained ResNet-50 using ASWL following the same training setting in \cite{ResNet} for ResNet-50. The hyper-parameter $\gamma$ was set at 0.5. Additionally, the pruning factor was set to 1. The models with the best results are selected.

Table \ref{table.ImageNet.Compare} compares ASWL with state-of-the-art pruning methods on ResNet-50. Through simultaneous training with layer-wise pruning from randomly initialized weights, ASWL achieves the top-1 accuracy of $76.5\%$ with a great pruning ratio ($86.1\%$). Note that this accuracy is higher than the uncompressed ResNet-50 baseline model ($76.1\%$). Considering the balance between the top-1 accuracy and the pruning ratio, ASWL provides a superior pruning result when compared to all existing pruning methods. If we consider accuracy alone, it is the second best. 

\section{Conclusion}
\label{sec.Conclusion}
In this paper, we proposed a novel pruning pipeline, Attention-based Simultaneous sparse structure and weight Learning (ASWL). In ASWL, we first use the attention mechanism to learn the importance of each layer in a network and determine the corresponding pruning ratio. Then, the layer-wise sparsity and weights are jointly learned from scratch in a unified training procedure. Our extensive experiments on benchmark datasets show that ASWL achieves outstanding pruning results in terms of accuracy, pruning ratio, and operating efficiency.
\clearpage

{\small \balance
\bibliographystyle{unsrtnat}
\bibliography{references}
}

\end{document}